\documentclass[letterpaper, 10 pt, conference]{ieeeconf}  

\IEEEoverridecommandlockouts                              

\usepackage{color}
\usepackage{url}
\usepackage{indentfirst}
\usepackage{amsmath}
\usepackage{amstext}
\usepackage{graphicx}
\usepackage{booktabs}
\newcommand{\tabincell}[2]{\begin{tabular}{@{}#1@{}}#2\end{tabular}}

\title{\LARGE \bf
Graph Convolution-Based Deep Reinforcement Learning for Multi-Agent Decision-Making in Mixed Traffic Environments
}

\author{Qi Liu$^{*}$, Zirui Li$^{*}$, Xueyuan Li, Jingda Wu, Shihua Yuan
\thanks{$^{*}$Qi Liu and Zirui Li contributed equally to this paper}
\thanks{Qi Liu, Zirui Li, Xueyuan Li, and Shihua Yuan are with the School of Mechanical Engineering, Beijing Institute of Technology, Beijing, China. (E-mails: 3120195257@bit.edu.cn; z.li@bit.edu.cn; lixueyuan@bit.edu.cn; yuanshihua@bit.edu.cn.)}
\thanks{Zirui Li is also with Department of Transport and Planning, Faculty of Civil Engineering and Geosciences, Delft University of Technology, Stevinweg 1, 2628 CN Delft, The Netherlands.}
\thanks{Jingda Wu is with the School of Mechanical and Aerospace Engineering, Nanyang Technological University, Singapore, 639798. (E-mail:  jingda001@e.ntu.edu.sg)}
}

\begin{document}

\maketitle
\thispagestyle{empty}
\pagestyle{empty}

\begin{abstract}

An efficient and reliable multi-agent decision-making system is highly demanded for the safe and efficient operation of connected autonomous vehicles in intelligent transportation systems. Current researches mainly focus on the Deep Reinforcement Learning (DRL) methods. However, utilizing DRL methods in interactive traffic scenarios is hard to represent the mutual effects between different vehicles and model the dynamic traffic environments due to the lack of interactive information in the representation of the environments, which results in low accuracy of cooperative decisions generation. To tackle these difficulties, this research proposes a framework to enable different Graph Reinforcement Learning (GRL) methods for decision-making, and compares their performance in interactive driving scenarios. GRL methods combinate the Graph Neural Network (GNN) and DRL to achieve the better decisions generation in interactive scenarios of autonomous vehicles, where the features of interactive scenarios are extracted by the GNN, and cooperative behaviors are generated by DRL framework. Several GRL approaches are summarized and implemented in the proposed framework. To evaluate the performance of the proposed GRL methods, an interactive driving scenarios on highway with two ramps is constructed, and simulated experiment in the SUMO platform is carried out to evaluate the performance of different GRL approaches. Finally, results are analyzed in multiple perspectives and dimensions to compare the characteristic of different GRL approaches in intelligent transportation scenarios. Results show that the implementation of GNN can well represents the interaction between vehicles, and the combination of GNN and DRL is able to improve the 
performance of the generation of lane-change behaviors. The source code of our work can be found at \url{https://github.com/Jacklinkk/TorchGRL}.

\end{abstract}

\section{INTRODUCTION}

Autonomous vehicles operate in highly complex driving environments. In an interactive traffic scenario, the driving environment is highly cooperative and dynamic, and the mutual effect between different traffic participants has a great influence on decision-making.\cite{Intro1}. It is critically important for each autonomous vehicle in interactive traffic scenarios to generate appropriate and cooperative behaviors. Therefore, intelligent multi-agent decision-making technology is highly demanded for autonomous vehicles to efficiently handle complex road environments and multi-agent interactions \cite{Intro2}.
 
The detailed overviews of decision-making technology for autonomous vehicles are presented in \cite{overview1, overview2}. In general, current technologies mainly focus on reinforcement learning (RL) methods due to the high complexity of the driving environments and the frequency of interaction with different agents \cite{Intro3}. The keys of reinforcement learning in interactive traffic scenarios can be summarized in two following aspects: 1) The accurate representation of the feature of each agent and the interaction between different 
vehicles. 2) The efficient modeling of the driving policy to generate reasonable behaviors \cite{Intro4}. Considering that RL method with manually designed policy model are easily affected by prior knowledge, which suffers from several weaknesses, including accuracy of generated behavior in interactive environments and generality between different scenarios. With the rapid development of deep learning methods in the field of supervised learning, incorporating neural networks into RL frameworks had shown large potential to improve the performance of RL method in recent researches. Thus, deep reinforcement learning (DRL) methods have been proposed to better ensure high efficient exploration and learning process of each agent \cite{Intro5}.

There have been some studies on the decision-making in interactive traffic scenarios with DRL methods \cite{qiao2018pomdp, bernhard2019addressing, hoel2020reinforcement}. However, such approaches only consider the individual features of each vehicle as the input of DRL, ignoring the mutual effects between pairs of vehicles. This will result in generating low cooperative behaviors in interactive traffic scenarios, which lead to danger or even traffic accident. Graph representation can accurately describe the mutual effects between pairs of agents, which enables modeling the relationship of vehicles in interactive traffic scenarios. Thus, some researches focus on GRL methods to model the interaction with graph representation. GRL methods is a combination of GNN and DRL to achieve the better decisions generation in interactive scenarios, the features of interactive scenarios are proceed by GNN, and cooperative behaviors are generated by DRL framework. In \cite{Intro6}, graph convolutional reinforcement learning method is proposed for multi-agent decision-making; two multi-head attention graph convolutional layers are utilized for features extraction of agents, and behaviors are generated by DQN for different agents. Experiment is carried out on the "MAgent" platform, and results show that the proposed methods substantially outperform existing methods in a variety of cooperative scenarios. In \cite{Intro7}, GNN and DQN are combined (named GCQ) for multi-agent cooperative controlling of CAVs in merging scenarios. Specifically, GNN is proposed to aggregate the information acquired from collaborative sensing, while cooperative lane-changing decisions are generated from DQN. Experiment is carried out in the SUMO platform, and results show good performance compared with the rule-based method and LSTM-based Q-learning method.

\begin{figure*}[thpb]
  \centering
  \includegraphics[scale=0.17]{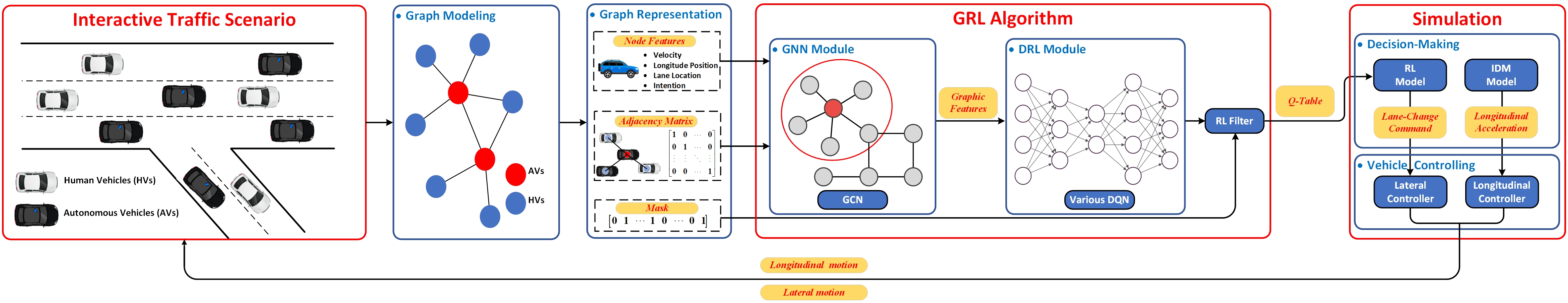}
  \caption{The schematic diagram of the proposed framework.}
  \label{figurelabel_1}
\end{figure*}

This paper proposes an innovative modular framework to analyze the performance of different GRL methods in interactive traffic scenarios.The traffic scenarios adopted in our work is constructed based on \cite{Intro7}. The designed algorithm in this paper is based on GRL, which consists of two modules: GNN and DRL. Specifically, several types of Deep Q-Learning methods are utilized as the DRL module, including DQN, Double DQN, Dueling DQN and Dueling-Double DQN. The graph representation of the multi-agent, including node feature matrix, adjacency matrix, and RL-filter, are selected as input, while the decision-making behaviors of each agent are generated as output. The general framework of our approach is modularized developed based on python, different categories of GNN and DRL algorithms can be replaced according to actual needs. Finally, simulation is carried out based on the SUMO platform \cite{dlr127994}, and the results are discussed in detail from multiple perspectives and dimensions.The schematic diagram of the designed framework is shown in Fig.\ref{figurelabel_1}.

The main contributions of this research are summarized as follows:

(1) An innovative modular framework is proposed to analyze the performance of GRL methods in decision-making in interactive traffic scenarios. This framework enables the combination of different types of GNN and DRL methods; and it can also be verified in various type of interactive traffic scenarios.

(2)The features of the multi-vehicle interactive environments are modeling by graphs and input into GNN, and cooperative behaviors are generated by combining different DRL methods.

\section{Problem Formulation}

The purpose of our work is to generate cooperative lane-change decisions for autonomous vehicles in an Interactive traffic scenario. To achieve this goal, a framework implemented with several GRL algorithms is proposed, which consists of three key components: Interactive Traffic Scenario, GRL Algorithm, and Simulation. The vehicles in the scenario are modeled as a graph, where nodes represent different vehicles and edges represent the interplay between every two vehicles. Graph representation is then processed into node features, adjacency matrix, and RL-filter. 
\begin{figure}[thpb]
  \centering
  \includegraphics[scale=0.42]{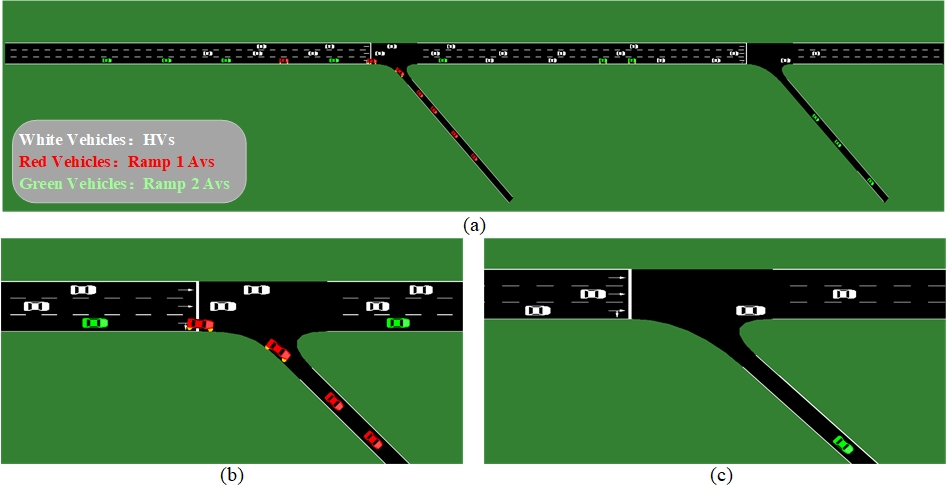}
  \caption{The traffic scenario constructed in the proposed research. (a) is the whole view of the traffic scenario; (b) is the view of ramp 1 where red vehicles drive out; (c) is the view of ramp 2 where green vehicles drive out.}
  \label{figurelabel_2}
\end{figure}The GRL Algorithm is
developed to train lane-changing policy, which takes graph representation as input and generates the Q values \cite{watkins1989learning} of different lane-change actions. The simulation part takes Q values as input, motions of each vehicle are generated to update the Interactive Traffic Scenario, thus enabling the continuous training of the GRL network.

The scenario in our research is constructed based on \cite{Intro7}, which is shown in Fig.\ref{figurelabel_2}. The scenario includes a 3-lane highway with two ramp exits. Two types of vehicles operate in this scenario: human-driven vehicles (HVs) and autonomous vehicles (AVs). Different vehicles have different driving tasks, and they need to cooperate to complete the scheduled driving tasks more efficiently. White vehicles represent HVs, entering from the left side of the highway and exiting from the right side. Colored vehicles represent AVs and enter from the left side of the freeway, specifically red vehicles exiting at ramp 1 and green vehicles exiting at ramp 2.

AVs interact with the traffic environment by taking lane-change actions at discrete time steps in the constructed traffic scenario. Upon taking the actions, the state of the traffic environment changes, and AVs receive a reward. Nevertheless, the environmental information can be completely observed within the observation range of AVs. Thus, this process can be modeled as Markov Decision Process (MDPs). For this constructed traffic scenario, four components define a finite horizon MDP: state space \((S)\), action apace \((a)\), reward function (R), and the discount factor \((\gamma)\) \cite{sigaud2013markov}.


\subsection{Input Representation}

The scenario is modeled as an undirected graph. Each vehicle in the scenario is regarded as the node of the graph, and the interaction between vehicles is regarded as the edge of the graph. More formally, the constructed graph is described as \(G=\{N, E\}\), where \(N=\{n_{i}, i\in\{1,2,...n\} \}\) is a set of node attributes and \(E=\{e_{ij}, i,j\in\{1,2,...n\} \}\) is a set of edge attributes. Specifically, \(n\) denotes the number of nodes in the constructed graph that is equal to the total number of vehicles. Then, graph representation is utilized to model the characteristics and interaction of vehicles. 
the state space is discrete and can be described as follows: 
\begin{equation}
S_{t}=[N_{t},A_{t},F_{t}]
\end{equation}

\noindent where \(N_{t}\in \mathbf{R}^{n\times f}\) denotes the node features matrix, \(A_{t}\in \mathbf{R}^{n\times n}\) denotes the adjacency matrix and \(F_{t}\in \mathbf{R}^{1\times n}\) denotes the RL-filter matrix. The above three matrices together constitute the state space matrix, each of the matrix is manipulated as follows. 


\subsubsection{Node Features Matrix}

Node features matrix represents the features of the constructed scenario. It consists of the feature matrix of each vehicle, which can be described as follows: 

\begin{equation}
N_{t}=
\begin{bmatrix}
 [V_{1},X_{1},L_{1},I_{1}]\\ 
 [V_{2},X_{2},L_{2},I_{2}]\\ 
 \cdots\\ 
 [V_{i},X_{i},L_{i},I_{i}]\\ 
 \cdots\\ 
 [V_{n},X_{n},L_{n},I_{n}] 
\end{bmatrix}
\end{equation}

\noindent where \([V_{i},X_{i},L_{i},I_{i}]\) represent the normalized state matrix of each vehicle. Specifically in the lane-change task, \(V_{i}=V_{i\_actual}/V_{max}\) denoted the normalized longitudinal speed of vehicles relative to the maximum longitudinal speed; \(X_{i}=X_{i\_longitudinal}/L_{highway}\) denotes the normalized longitudinal coordinate of vehicles relative to the length of highway; \(L_{i}\) denotes the one-hot encoding matrix of current lane position (left-lane, right-lane and middle-lane) of vehicles; \(I_{i}\) denotes the 
one-hot encoding matrix of current intention of (change to left-lane, change to right-lane and go straight) of vehicles. 


\subsubsection{Adjacency Matrix}

Adjacency matrix represents the interaction between vehicles, which is embodied in the information sharing between vehicles in the constructed scenario. The adjacency matrix is calculated based on the following assumption: 1) All vehicles can share information with themselves. 2) All AVs can share information. 3) All AVs can share information with HVs, which are in AVs' sensing range. The derivation of the adjacency matrix is as follows:

\begin{equation}
A_{t}=
\begin{bmatrix}
 e_{11}& e_{12}& \cdots&  & \cdots& e_{1n}\\ 
 e_{21}& e_{22}& \cdots&  & \cdots& e_{2n}\\ 
 \vdots& \vdots&  \ddots&  &  & \vdots\\ 
 & &  &  e_{ij}&  & \\ 
 \vdots& \vdots&  &  &  \ddots& \vdots\\ 
 e_{n1}& e_{n2}& \cdots&  & \cdots& e_{nn} 
\end{bmatrix}
\end{equation}

\noindent where \(e_{ij}\) denotes the edge value of \(\textrm{vehicle}_i\) and \(\textrm{vehicle}_j\). \(e_{ij}=1\) when \(\textrm{vehicle}_i\) and \(\textrm{vehicle}_j\) share information; while \(e_{ij}=0\) when \(\textrm{vehicle}_i\) and \(\textrm{vehicle}_j\) share no information.



\subsubsection{RL-filter Matrix}

The construction of the RL-filter matrix is to filter out the corresponding terms of HVs for the output of the GRL algorithm. The derivation of the RL-filter matrix is as follows:

\begin{equation}
    F_{t}=[f_{1},f_{2},\cdots,f_{i},\cdots,f_{n}]
\end{equation}

\noindent where \(f_{i}=0\ \textrm{or}\ 1\) . If \(vehicle_i\) belongs to AVs controlled by GRL algorithm, \(f_{i}=1\); otherwise \(f_{i}=0\).

\subsection{Action Space}

The action space is discrete. At each time step, the action space of AVs in the scenario is composed of different lane changing instructions that can be described as follows:

\begin{equation}
    a=[change\ to\ left, go\ straight, change\ to\ right]
\end{equation}

\subsection{Reward Function}

The details of the reward function can refer to \cite{Intro6}. In general, the reward function consists of four parts: intention reward, average speed reward, lane-changing penalty, and collision penalty. The reward function is defined as follows:

\begin{equation}
    R=w_{1}R_{\textrm{I}}+w_{2}R_{\textrm{AS}}+w_{3}P_{\textrm{LC}}+w_{4}P_{\textrm{C}}
\end{equation}
where \(w_{i}\) denotes the weight of each item; \(R_{\textrm{I}}\) denotes the intention reward, \(R_{\textrm{AS}}\) denotes the average speed reward; \(P_{\textrm{LC}}\) denotes the lane-changing penalty, \(P_{\textrm{C}}\) denotes the collision penalty.

\section{GRL Algorithm}

\begin{figure*}[thpb]
  \centering
  \includegraphics[scale=0.25]{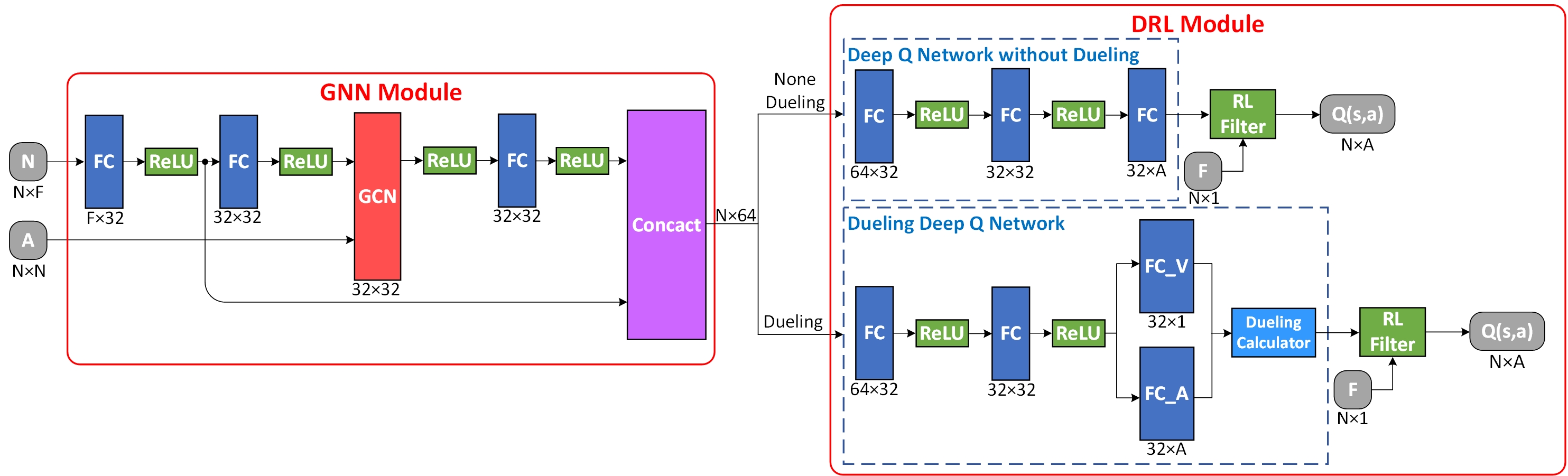}
  \caption{The network structures of GRL algorithm.}
  \label{figurelabel_3}
\end{figure*}

The GRL algorithm is the key part of the proposed framework, consisting of GNN and DRL. Based on the graph representation of scenario and reward functions, GRL can explore optimized lane-change strategies to ensure autonomous vehicles' efficient and safe completion of driving tasks. The detailed algorithm design of GNN and DRL module will be further illustrated. The detailed network structures of GRL is shown in the Fig.\ref{figurelabel_3}.

\subsection{GNN Module}

The GNN module consists of fully connected layers (FC) and graph convolutional network (GCN). GCN is a type of GNN, that extracts features of graph-structured data based on an efficient variant of convolutional neural networks \cite{kipf2016semi}. Before implementing the graph convolution process, the node features matrix \(N_{t}\) is firstly input into the FC to map the node feature to sample markup space, which is described as follows:

\begin{equation}
    N_{\textrm{FC}}=\phi^{\textrm{FC}}(N_{t})
\end{equation}

\noindent where \(N_{\textrm{FC}}\) denotes the node feature matrix proceed by the FC; \(\phi^{\textrm{FC}}\) denotes the neural network with FC.

Then, \(N_{\textrm{FC}}\) and adjacency matrix \(A_{t}\) are firstly input into the GCN to generate graphical convolution feature; and the graph convolution process can be expressed as following formulation \cite{kipf2016semi}:

\begin{equation}
    G_{t}=\phi^{\textrm{GCN}}(N_{\textrm{FC}}, A_{t})=\sigma(D_{t}^{\frac{1}{2}}A_{t}D_{t}^{-\frac{1}{2}}N_{\textrm{FC}}W_{t}+b)
\end{equation}

\noindent where \(G_{t}\) denotes the graph convolutional features proceed from GCN; \(\phi^{\textrm{GCN}}\) denotes the graph convolution operator; matrix \(D_{t}\) is computed based on \(A_{t}\), specifically, \(D_{ii}=\Sigma_{j}A_{ij}\); \(W_{t}\) denotes a layer-specific trainable weight matrix; \(b\) denotes the offset, specifically, \(b=0\) is defined in this research; \(\sigma\) denotes an activation function, specifically, ReLU \cite{glorot2011deep} is chosen in this research. 

Graph convolutional features are then input into the DRL module, then the Q value of lane changing actions for AVs is generated by filtering through the RL-filter matrix. The derivation of the Q value is as follows:

\begin{equation}
    Q(s,a)=F_{t}\cdot[\phi^{\textrm{DRL}}(G_{t})]
\end{equation}

\noindent where \(Q(s,a)\) denotes the Q value of lane changing actions for AVs; \(\phi^{DRL}\) denotes the policy neural network.

\subsection{DRL Module}

This part elaborates on the principles and implementation process of different DRL algorithms carried out in this research. The comparison of different DRL algorithms is shown in the Table.


\subsubsection{Deep Q-Network (DQN)}

Q-learning is one of the most representative value-based reinforcement learning methods based on the estimation of Q value \cite{watkins1989learning}. Q value can be expressed as \(Q(s,a)\), defined as the expectation of future rewards when taking action \(\{a,a\in A\}\) at state \(\{s,s\in S\}\) at a certain time step; in addition, reward \(r\) is generated to the agent according to the response of the environment. The main idea of Q-learning is selecting the action that can obtain the greatest reward based on the Q value; then, Q value is updated as follows:

\begin{equation}
    Y_{t}^{Q}=R_{t+1}+\gamma\  \underset{a}{\textrm{argmax}}Q(s_{t+1},a)
\end{equation}

\noindent where \(\gamma\) denotes the attenuation factor of the reinforcement learning process.

However, Q-learning suffers from the curse of dimension. When the dimension of the state space is too large, the construction, searching, and updating of Q value will cause high computational complexity; thus, the efficiency of the algorithm will be greatly reduced, or even become impossible to solve the reinforcement learning problem. With the development of deep learning technology, DQN was proposed in \cite{mnih2013playing}, a neural network is constructed to estimate the Q value of the action space to extend the state space to high dimensions. The online Q network can be parametrized as \(Q(s,a;\theta)\). The state space of environment is chosen as the input of the neural network; while the Q value of the action space is generate as the output of the neural network, and finally the next action of the agent is determined. 

DQN includes the following elements: replay buffer \(D\) and a fixed target Q network parametrized as \(Q(s,a;\theta^{\prime})\). During the learning process of the agent, transition \((s,a,r,s^{\prime})\) is stored in the replay buffer and sampled at each time step. The calculation process of target Q value can be expressed by the following formula:

\begin{equation}
    Y_{t}^{\textrm{DQN}}=R_{t+1}+\gamma\  \underset{a}{\textrm{argmax}}Q(s_{t+1},a;\theta^{\prime})
\end{equation}

The output value of the target Q network \(y\) is compared with the estimated value of the online Q network to calculate the loss, a gradient descent step on \((Y_{i}^{\textrm{DQN}}-Q(s_{i},a_{i};\theta))^{2}\) is carried out to update \(\theta\). In addition, \(\theta\) is synchronized to \(\theta^{\prime}\) every certain number of iterations.

DQN also suffers from several weaknesses: 1) The target Q value is overestimated due to the max operator. 2) The online Q network has the same structure as the target Q network, resulting in inefficient learning and convergence.


\subsubsection{Double DQN}

Double DQN is proposed to solve the overestimation problem of DQN \cite{van2016deep}. Double DQN and DQN lie in the different estimation methods of target Q value. The main idea is to construct different functions for action selection and value evaluation. When updating the target Q value, the subsequent actions are selected from the online Q network, while the assessment of the Q value depends on the target Q network. The estimation of the target Q value of Double DQN is derived as follows:

\begin{equation}
    Y_{t}^{\textrm{Double}}=R_{t+1}+\gamma\ Q(s_{t+1}, \underset{a}{\textrm{argmax}}Q(s_{t+1},a;\theta);\theta^{\prime})
\end{equation}


\subsubsection{Dueling DQN}

Dueling DQN is proposed to optimize the network structures of DQN \cite{wang2016dueling}. Specifically, Dueling DQN consists of two separate estimators: one for the state value function, defined as \(V(s;\theta,\beta)\); another for the state-dependent action advantage function, defined as \(A(s,a;\theta,\alpha)\). Outputs from the two estimators are finally combined to generate the target Q value \(Q(s,a;\theta,\alpha,\beta)\). The main benefit of this factoring is to generalize learning across actions without imposing any change to the underlying reinforcement learning algorithm to ensure better policy evaluation. The estimation of the Q function of Dueling DQN is derived as follows:

\begin{equation}
\begin{split}
    Q(s,a;\theta,\alpha,\beta)=V(s;\theta,\beta)+
    (A(s,a;\theta,\alpha)-\\
    \frac{1}{\left|\mathcal{A}\right|}\underset{a^{\prime}}{\sum}A(s,a^{\prime};\theta,\alpha))
\end{split}
\end{equation}

\noindent where \(\theta\) denotes the parameters of the online Q network, while \(\alpha\) and \(\beta\) denotes the parameters of the two streams of fully-connected layers; \(\left|\mathcal{A}\right|\) denotes the absolute value of the average of \(A(s,a;\theta,\alpha)\)

Thus, the target Q value of Dueling DQN can be calculated as follows:

\begin{equation}
    Y_{t}^{\textrm{Dueling}}=R_{t+1}+\gamma\  \underset{a}{\textrm{argmax}}Q(s_{t+1},a;\theta^{\prime},\alpha^{\prime},\beta^{\prime})
\end{equation}


\subsubsection{Dueling Double DQN (D3QN)}

D3QN combines the improvements from Double DQN and Dueling DQN to optimize the estimation of Q value while ensuring better strategy evaluation. The estimation of target Q value is derives as follows:

\begin{equation}
\begin{split}
    & Y_{t}^\textrm{{D3QN}}=R_{t+1}+\\
    & \hspace{5mm} \gamma\ Q(s_{t+1}, \underset{a}{\textrm{argmax}}Q(s_{t+1},a;\theta,\alpha,\beta);\theta^{\prime},\alpha^{\prime},\beta^{\prime})
\end{split}
\end{equation}

\section{Experiment}

In this section, the construction of simulation environment is described, and implementation details of the proposed research is presented.

\subsection{Simulation Environment}

\begin{figure}[thpb]
  \centering
  \includegraphics[scale=0.4]{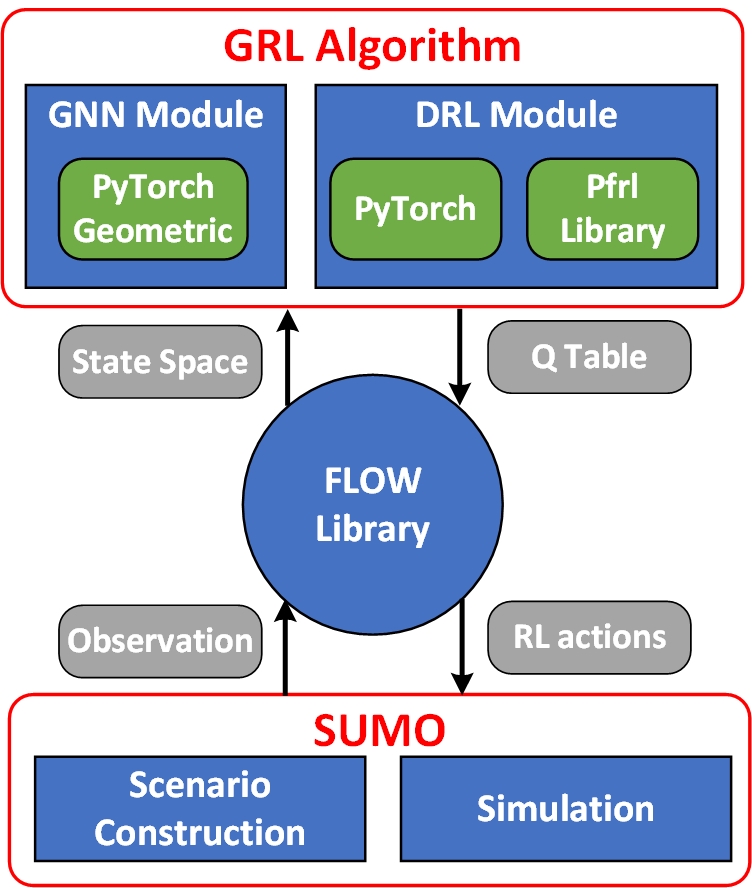}
  \caption{The program structure of the proposed framework}
  \label{figurelabel_4}
\end{figure}

The proposed framework is developed based on python and various third-party libraries. The program structure is shown in Fig.\ref{figurelabel_4}, and the information of implemented third-party libraries is presented in Table \ref{table_1}.

\begin{table}[h]
\caption{The information of implemented third-party libraries}
\label{table_1}
\begin{center}
\begin{tabular}{ccc}
\toprule
Library & Refs & Function\\
\midrule 
Pytorch & \cite{paszke2019pytorch} & The core framework of neural network\\
Pfrl & \cite{JMLR} & The DRL library based on Pytorch\\
\tabincell{c}{Pytorch\\Geometric} & \cite{Fey/Lenssen/2019} & The GNN library based on Pytorch\\
\tabincell{c}{FLOW\\Library} & \cite{wu2017flow} & \tabincell{c}{The interface between traffic simulators\\ and DRL libraries}\\
\bottomrule
\end{tabular}
\end{center}
\end{table}

The traffic scenario and experiments are constructed based on the SUMO platform. It consists of a lateral control module and longitudinal control module. Q value of different lane-change actions from the GRL algorithm is input into the lateral control module, and control commands are finally generated to update the status of each vehicle, thus enabling the continuous training of the GRL algorithm. The two control modules are described in detail as follows:


\subsubsection{Longitudinal Control Module}

Acceleration commands are generated from the longitudinal control module. The longitudinal control of both HVs and AVs is achieved by the IDM model \cite{treiber2013traffic} embedded in SUMO. 


\subsubsection{Lateral Control Module}

Lang-change commands are generated from the lateral control module. The lateral control of HVs is achieved by
the LC2013 lane-changing model \cite{erdmann2015sumo} embedded in SUMO, while the lateral control of AVs is achieved by the proposed GRL algorithm.

\subsection{Implementation Details}

Simulation parameters include the parameters of traffic scenario that are summarized in Table \ref{table_2}, and the parameters of GRL algorithm that are summarized in Table \ref{table_3}. 
\begin{table}[h]
\caption{Parameters of traffic scenario}
\label{table_2}
\begin{center}
\begin{tabular}{ccc}
\toprule
Parameters & Symbols & Value\\
\midrule 
Number of HVs & \(N_{\textrm{HVs}}\) & 20\\
Number of AVs & \(N_{\textrm{AVs}}\) & 20\\
Number of AVs driven out by ramp 1 & \(N_{\textrm{r1}}\) & 10\\
Number of AVs driven out by ramp 2 & \(N_{\textrm{r2}}\) & 10\\
Length of Highway & \(L\) & 500m\\
Longitudinal position of ramp 1 & \(X_{\textrm{r1}}\) & 200m\\
Longitudinal position of ramp 2 & \(X_{\textrm{r2}}\) & 400m\\
Speed limit for HVs & \(V_\textrm{{maxHVs}}\) & 60km/h\\
Speed limit for AVs & \(V_\textrm{{maxAVs}}\) & 75km/h\\
Inflow of HVs & \(P_\textrm{{HVs}}\) & 0.3veh/s\\
Inflow of AVs & \(P_\textrm{{AVs}}\) & 0.15veh/s\\
\bottomrule
\end{tabular}
\end{center}
\end{table}

\begin{table}[h]
\caption{Parameters of GRL algorithms}
\label{table_3}
\begin{center}
\begin{tabular}{ccc}
\toprule
Parameters & Symbols & Value\\
\midrule 
Number of training episode & \(N_{\textrm{training}}\) & 150\\
Step size of the random exploration phase & \(S_{\textrm{random}}\) & 20000\\
Batch size & \(M_{\textrm{batch}}\) & 32\\
Replay buffer capacity & \(M_{\textrm{replay}}\) & \(10^{6}\)\\
Discount factor & \(\gamma\) & 0.9\\
Optimizer & - & Adam\\
Learning rate & \(\eta\) & \(10^{-4}\)\\
Online network update frequency & \(N_{\textrm{online}}\) & 10\\
Target network update frequency & \(N_{\textrm{target}}\) & 1000\\
Target network update method & - & soft\\
Soft target network update rate & \(\eta_{soft}\) & 0.01\\

\bottomrule
\end{tabular}
\end{center}
\end{table}

Different exploration strategies are adopted at different training stages. Define \(\pi(s)\) as the lane-changing strategy adopted by the AVs in state \(s\). At the beginning, a random exploration phase with a specific step is carried out to expand the exploration space of AVs. During the random exploration phase, AVs are sampled in the action space to generate random lane-change behavior, defined as \(\pi(s)=\textrm{random}(a)\). After the random exploration phase, the epsilon greedy strategy is used to generate the lane-change behavior of the AVs according to the Q table calculated from the GRL algorithm, and network parameters are updated at regular step intervals. The derivation of epsilon greedy strategy is as follows:

\begin{equation}
\pi(s)=
\left\{
\begin{array}{ll}
\textrm{random}(a) & P=\epsilon\\
\textrm{argmax}Q(s_{t},a) &{ P=1-\epsilon}
\end{array}\right.
\end{equation}

\noindent During the testing process, the greedy strategy is used to generate lane-change behavior, defined as \(\pi(s)=\textrm{argmax}Q(s_{t},a)\).

\subsection{Results}

Experimental results are presented in this part, including reward, loss, and average Q value of different GRL algorithms. The rule-based method is selected as the baseline, and LC2013 lane-changing model is applied to AVs instead of the controlling of GRL algorithm. 

\subsubsection{Reward}

\begin{figure}[thpb]
  \centering
  \includegraphics[scale=0.5]{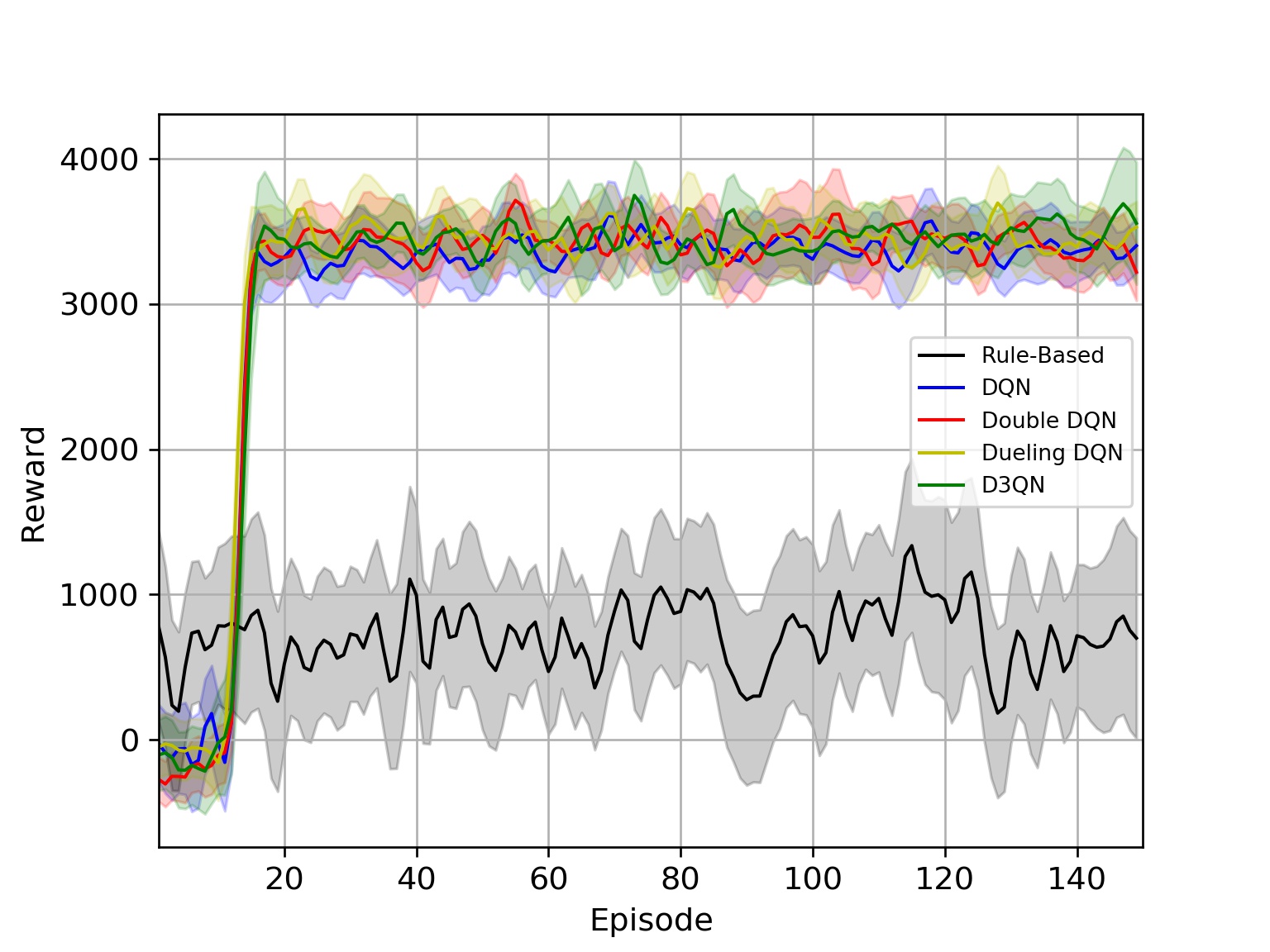}
  \caption{The reward curve of four GRL algorithms. The
shaded areas show the standard deviation for 3 random seeds.}
  \label{figurelabel_5}
\end{figure}

During the training process, the reward for each training episode is shown in Fig.\ref{figurelabel_5}. The average reward from the end of random exploration phase is summarized in Table \ref{table_4}.

\begin{table}[h]
\caption{Average training reward of four GRL algorithms}
\label{table_4}
\begin{center}
\begin{tabular}{cc}
\toprule
Algorithm & Average Training Reward\\
\midrule 
DQN & 3375.50\\
Double DQN & 3433.23\\
Dueling DQN & 3453.67\\
D3QN & 3460.90\\
\bottomrule
\end{tabular}
\end{center}
\end{table}

\subsubsection{Loss}

\begin{figure}[thpb]
  \centering
  \includegraphics[scale=0.5]{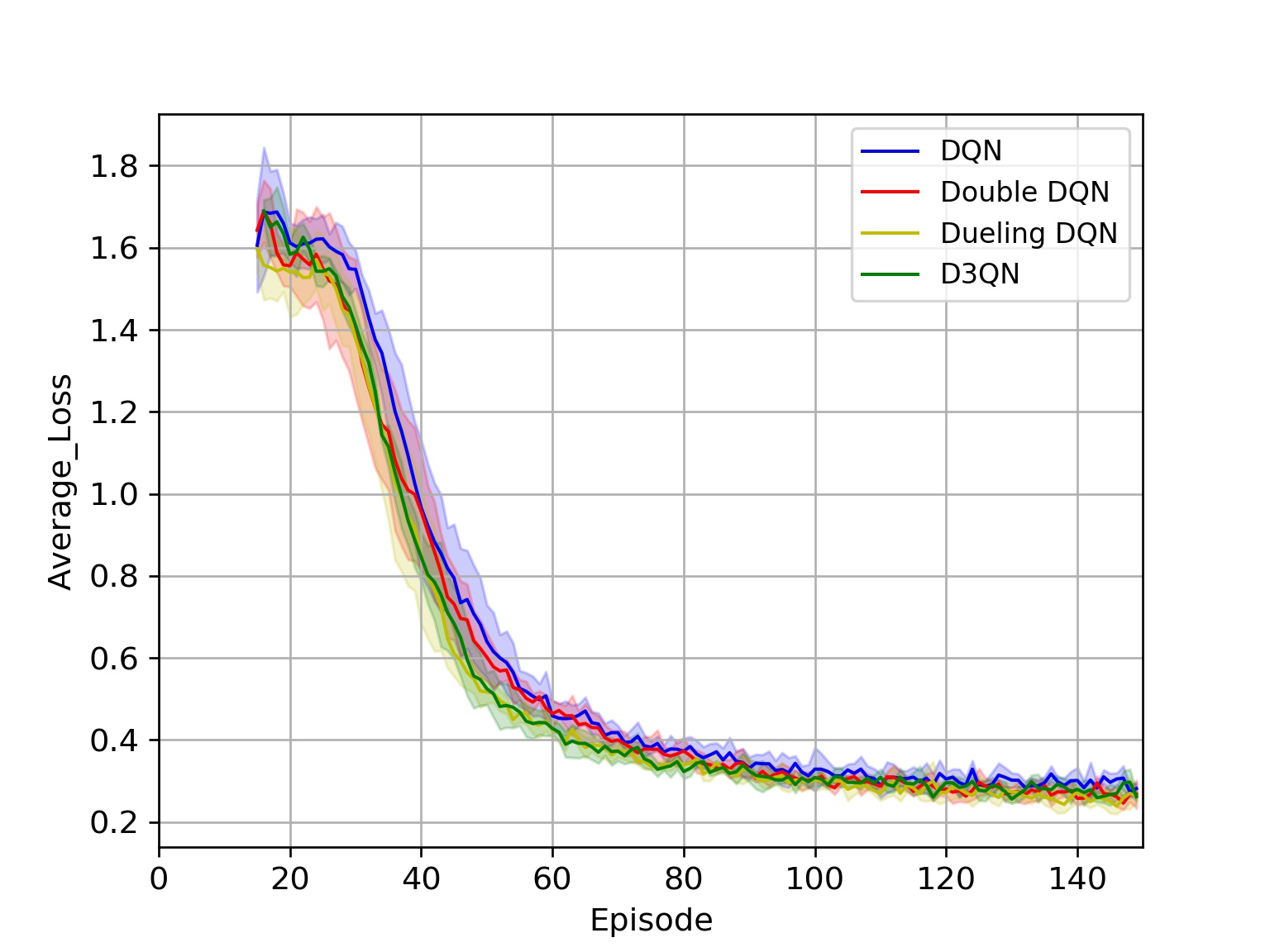}
  \caption{The loss curve of four GRL algorithms. The
shaded areas show the standard deviation for 3 random seeds.}
  \label{figurelabel_6}
\end{figure}

During the training process, the loss is calculated for each training episode, which is defined as. The loss for each training episode is shown in Fig.\ref{figurelabel_6}, and the average loss in the whole training process is summarized in Table \ref{table_5}.


\subsubsection{Average Q Value}

\begin{figure}[thpb]
  \centering
  \includegraphics[scale=0.5]{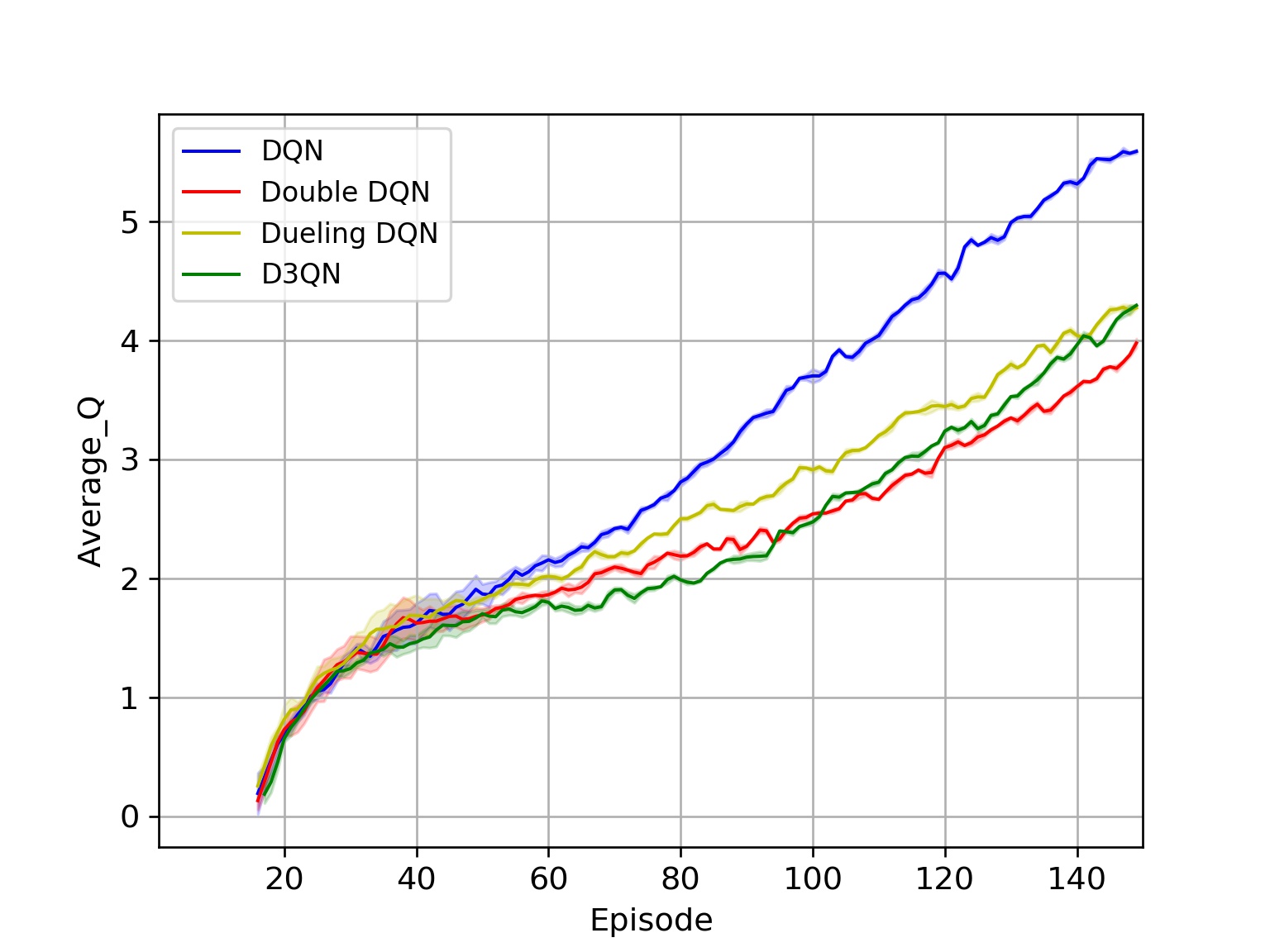}
  \caption{The average Q value curve of four GRL algorithms. The
shaded areas show the standard deviation for 3 random seeds.}
  \label{figurelabel_7}
\end{figure}

During the training process, the average of Q values in the Q table generated by the GRL network is calculated for each training episode to compare the estimation effects on the Q value of different algorithms. The average Q value for each training episode is shown in Fig.\ref{figurelabel_7}.

\subsection{Analysis}

In this part, experimental results are presented and analyzed to compare the performance of different GRL algorithms applied in our proposed framework. 

As shown in Fig.\ref{figurelabel_5}. The rewards of the four GRL algorithms are much higher than the rule-based method. It proves that the implementation of GNN can well represents the interaction between vehicles, and the combination of GNN and DRL is able to improve the performance of the generation of lane-change behaviors. In the random exploration phase, due to the random behavior generation of AVs, the reward is low with severe fluctuation; after the random exploration phase, the reward rises rapidly, and fluctuation becomes steady. However, the reward curve of the four algorithms didn’t show significant differences. It can be concluded from Table that the average reward of Double DQN is higher than that of DQN; and the average reward of Dueling DQN and D3QN is much higher than that of DQN. This is because the establishment of Dueling Network can better optimize the behavior generation of AVs.

As shown in Fig.\ref{figurelabel_6}. The loss of the four algorithms in the training process decreases with the increase of the training episode, and converges to a stable value. DQN and Double DQN have no obvious difference in the loss convergence process, but it is significantly higher than Dueling DQN and D3QN, and the convergence speed is relatively slow. After the loss convergence process, the loss curves of the four algorithms have no noticeable difference. It can be seen from the table that the average loss of Dueling DQN is the lowest, and there is no significant difference between the average loss of DQN and Double DQN. This result can prove that optimizing the strategy with Dueling Network can effectively reduce the loss in the training process.

It can be seen from Fig.\ref{figurelabel_7} that as the training episode increases, the average Q value increases continually; however, the variation trend of the average Q value for the four algorithms is quite different. The curve of DQN is significantly higher than the other three algorithms, but the average reward is relatively low; this is due to the problem of overestimation caused by the max operator when calculating the target Q value. The curve of Double DQN and D3QN changes relatively smoothly, which shows that the implementation of Double operation can effectively improve the effect of Q value evaluation. The average Q value of Dueling DQN is lower than DQN but higher than Double DQN and D3QN, indicating that the establishment of Dueling Network can also benefit the evaluation of Q value.

We then test the four GRL algorithms under the same experimental parameter settings for ten episodes. During the test process, we restricted the lane-changing behavior of all the vehicles to avoid collisions as much as possible. The average reward is calculated, and the results are shown in Table \ref{table_6}.

\begin{table}[h]
\caption{Average testing reward of four GRL algorithms}
\label{table_6}
\begin{center}
\begin{tabular}{cc}
\toprule
Algorithm & Average Testing Reward\\
\midrule 
DQN & 3246.10\\
Double DQN & 3283.57\\
Dueling DQN & 3295.97\\
D3QN & 3346.74\\
\bottomrule
\end{tabular}
\end{center}
\end{table}

Results show that D3QN has the highest reward, followed by Dueling DQN, Double DQN, and DQN. It can be concluded that the improved DQN methods have shown promising results in the constructed traffic scenario. 

\section{Conclusion}
This research proposes a modular framework with high generality that consists of three key modules: traffic scenario, GRL algorithm, and simulation. The proposed framework can be used to verify the effects of different GRL methods in various traffic scenarios according to the actual demands. A 3-lane highway with two ramp exits scenario is constructed, and four different types of GRL algorithms are applied and evaluated in our framework. The experimental results show that the combination of GNN and DRL can well solve the decision-making task in an interactive 
traffic scenario. The designed framework can also realize the verification and comparison of different GRL algorithms in the field of the intelligent transportation system. Future work will focus on the designing of more discriminative traffic scenarios to better verify the effect of the algorithm, and the implementation of more types of GRL algorithms into the proposed framework. 

\bibliographystyle{IEEEtran}
\bibliography{myref}

\end{document}